\documentclass[11pt]{article}

\usepackage[utf8]{inputenc}
\usepackage[T1]{fontenc}
\usepackage{mathptmx}

\usepackage[margin=1in]{geometry}
\usepackage{setspace}
\usepackage{ragged2e}
\usepackage{enumitem}
\setlist{nosep,leftmargin=1.5em}

\usepackage{array}
\usepackage{tabularx}

\usepackage{listings}
\usepackage{xcolor}
\lstdefinestyle{jsonstyle}{
  basicstyle=\ttfamily\footnotesize,
  breaklines=true,
  breakatwhitespace=false,
  columns=fullflexible,
  frame=single,
  framesep=6pt,
  xleftmargin=1.5em,
  xrightmargin=1em,
  showstringspaces=false
}

\usepackage[hidelinks]{hyperref}
\usepackage{url}

\newcommand{\refentry}[1]{\par\noindent\hangindent=0.5in\hangafter=1 #1\par\vspace{0.5em}}

\usepackage{titlesec}
\titleformat{\section}{\normalfont\large\bfseries}{}{0pt}{}
\titleformat{\subsection}{\normalfont\normalsize\bfseries}{}{0pt}{}
\begin{document}

\begin{center}
{\large\bfseries A Rubric-Based Controlled Comparison of Frontier Language Models on Expert-Authored Clinical Reasoning Tasks\par}
\vspace{1.0em}
{\normalsize Dr.~Samiha A.~Ismail\textsuperscript{1}, Fan X.~Chen, Ph.D.\textsuperscript{2}, Ali Merali\textsuperscript{3}\par}
\vspace{0.6em}
{\small
\textsuperscript{1}Prime Analytics Consulting Limited\\
\textsuperscript{2}Talbert House; Thomas More University\\
\textsuperscript{3}MAKZ\par}
\end{center}
\vspace{0.5em}

\begin{center}\textbf{Abstract}\end{center}

Multiple-choice medical benchmarks are increasingly saturated, and recent rubric-based evaluations such as HealthBench have shown that open-ended clinical performance is far from solved -- its ``Hard'' subset top score remains 32\%. We present a small, deliberately difficult evaluation dataset of five clinician-authored clinical scenarios spanning four specialties (anaesthesia, internal/family medicine, emergency medicine, and obstetrics), each accompanied by an atomic, weighted, MECE rubric (25--62 criteria per task; 184 criteria total) authored from a clinician-drafted golden answer. We evaluate three frontier models: GPT 5.4, Claude Opus 4.7, and Gemini 3.1 Pro. Mean rubric pass rates were 0.47 (Claude), 0.38 (GPT), and 0.37 (Gemini). The central finding is an inversion of clinical priority: the highest-weighted (weight-5, critical) criteria passed at only 32.4--41.7\%, while low-stakes weight-1 criteria passed at 80--90\%. 55 of 108 critical (weight-5) criteria (51\%) were satisfied by no model. Three LLM autoraters reproduced expert met/not-met labels on 92.8--94.6\% of 552 graded criteria. We position this as a methods-and-preliminary-findings contribution: the five tasks demonstrate a scalable, defensible pipeline ready to develop into a large-scale benchmark.

\section{1.\quad Introduction}

Large language models now achieve near-ceiling accuracy on multiple-choice medical question-answering benchmarks. Med-PaLM 2 reached up to 86.5\% on the MedQA dataset, an improvement of roughly 19 percentage points over its predecessor (Singhal et al., 2025), and subsequent frontier models have pushed higher still. This saturation has shifted the field toward open-ended, rubric-based evaluation that better reflects real clinical work. HealthBench (Arora et al., 2025), built with 262 physicians who had practiced in 60 countries, evaluated models on 5,000 realistic health conversations scored against 48,562 unique rubric criteria. Crucially, HealthBench remains unsaturated: on its 1,000-example ``Hard'' subset, the strongest model evaluated achieved a score of 0.32, compared with 0.60 on HealthBench overall, leaving substantial headroom for the next generation of models (Arora et al., 2025). In parallel, the AMIE program has used OSCE-style, axis-based human and auto-rater evaluation to study diagnostic dialogue and longitudinal disease management (Tu et al., 2025).

Yet a specific capability gap remains under-measured: synthesis, prioritisation, and safety reasoning under contradictory evidence and resource constraints. Real clinical decisions require integrating signals that point in opposite directions (for example; having more than one clinical diagnosis which have conflicting treatments), recognising drug--disease and drug--drug interactions, anticipating downstream resource needs, and making time-bound commitments where the safe action is sometimes to \textit{withhold} an intervention. MCQ benchmarks cannot probe these behaviours, and even rubric-based conversational benchmarks tend to reward completeness and communication quality more than they penalise the omission of a single decisive, high-stakes inference.

We address this gap with a small but methodologically rigorous study. Our contributions are: (1)~\textbf{A clinician-authored evaluation methodology} in which practising clinicians write organic, narrative scenarios with synthetic patient data and draft an ideal golden answer from which atomic, weighted, MECE rubric criteria are constructed, (2)~\textbf{a controlled comparison} of three frontier models (GPT 5.4, Claude Opus 4.7, Gemini 3.1 Pro) under an identical, deliberately conservative single-turn harness, reported in the terminology of OpenAI's third-party-evaluation framework (claim type, harness, contamination, refusals, reward hacking); (3)~\textbf{a weight- and category-resolved failure analysis} that surfaces an inversion of clinical priority: models pass trivial criteria and fail critical ones, and we identify 55 of 108 critical (weight-5) criteria (51\%) that no frontier model satisfied; (4)~\textbf{an autorater calibration study} showing that three LLM graders reproduce expert met/not-met labels on 92.8--94.6\% of 552 criteria, and (5) demonstrate a scalable pipeline whose diagnostic value emerges at scale, ready to develop into a large-scale benchmark.

\section{2.\quad Related Work}

\textbf{\textit{Multiple-choice medical QA.}} MedQA (Jin et al., 2021) compiles USMLE-style questions; MedMCQA (Pal et al., 2022) and PubMedQA (Jin et al., 2019) extend coverage to Indian entrance exams and PubMed-abstract reasoning. MultiMedQA (Singhal et al., 2023) aggregated these with consumer-health queries and a human-evaluation framework, and the Med-PaLM models demonstrated rapid accuracy gains, ultimately reaching 86.5\% on MedQA (Singhal et al., 2025). These benchmarks measure recall and single-step reasoning but not synthesis under contradiction.

\textbf{\textit{Rubric-based and conversational evaluation.}} HealthBench (Arora et al., 2025) is the closest prior work: physician-written, conversation-specific rubrics scored by a model-based grader, with stratification by theme and behavioural axis (accuracy, completeness, instruction following, communication). HealthBench Consensus filters to 34 multiply-validated high-impact criteria, and HealthBench Hard isolates the cases on which frontier models score 32\%. The AMIE line of work (Tu et al., 2025; Saab et al., 2025; Palepu et al., 2025) evaluates diagnostic and management dialogue in randomized, blinded OSCE-style studies against primary care physicians, using clinically meaningful axes (history-taking, diagnostic accuracy, management reasoning, communication, empathy) and an LLM auto-rater whose ratings were reported to align well with human raters and to be comparable to inter-specialist agreement (Tu et al., 2025). MedHELM (Bedi et al., 2025) provides a clinician-validated taxonomy of 121 real-world tasks across 5 categories and 22 subcategories, evaluated with an LLM-jury that achieved an intraclass correlation of 0.47 against clinician ratings, marginally exceeding clinician--clinician agreement (0.43).

\textbf{\textit{Clinical reasoning benchmarks.}} ER-Reason (Mehandru et al., 2025) evaluates LLM reasoning across the emergency-department workflow using real longitudinal notes and Script-Concordance-style belief-updating questions scored against emergency-physician annotations. Two of its findings motivate our design: reasoning models do not reliably outperform non-reasoning ones, and accuracy degrades as evidence accumulates, plausibly due to error propagation from miscalibrated differential rankings at earlier timesteps (Mehandru et al., 2025). ER-Reason also reports a sobering inter-physician baseline of 52.6\% exact agreement on the final-diagnosis ordinal score, underscoring that raw agreement, not perfection, is the relevant yardstick for clinical graders.

\textbf{\textit{Rubric rewards.}} Recent work treats rubrics not only as evaluation instruments but as training signals: Gunjal et al. (2025) extend RL with verifiable rewards beyond verifiable domains using weighted rubric criteria, reporting substantial relative improvement on HealthBench over Likert-based reward baselines, and Liu et al. (2025) study scalable rubric generation.

In short, the current project is distinguished by the combination of (i) clinician-authored, golden-answer-derived rubrics, (ii) atomic weighted criteria tagged by clinical category, and (iii) a multi-model controlled comparisons under a fixed harness with explicit validity accounting.

\section{3.\quad Methods}

\subsection{3.1\quad Dataset construction}

Clinicians who actively practise in acute or primary care settings authored five organic, narrative clinical scenarios, each in their own specialty, using HIPAA-compliant synthetic patient data (if applicable). Every author is a senior front-line decision-maker who encounters these situations in their own clinical work: in practice, residents typically carry first-line decision-making responsibility and escalate to an attending for confirmation where needed, and the scenarios reflect the kinds of real-world cases the team manages directly. Some scenarios include attached input artefacts (e.g., a pre-anaesthetic assessment document). Each prompt specifies a clinically meaningful required output such as a ranked differential with reasoning columns, a structured table, a SOAP-style plan, or a set of time-bound commitments under resource constraints, and is designed to demand multi-step chained reasoning under contradictory anchors (e.g., an admission diagnosis contradicted by a later biomarker trend). The five tasks are summarised in Table 1.

\vspace{0.5em}
\noindent\textbf{Table 1. The five evaluation tasks.}
\vspace{0.3em}

\noindent\begin{tabularx}{\textwidth}{|c|>{\RaggedRight\arraybackslash}p{2.7cm}|>{\RaggedRight\arraybackslash}X|c|}
\hline
\textbf{Task} & \textbf{Specialty} & \textbf{Scenario} & \textbf{\# of Criteria} \\
\hline
1 & Anaesthetist & Intraoperative cardiac arrest during prone neurosurgery (cerebellar tumour resection) & 62 \\
\hline
2 & Family/Internal Medicine & Overnight digoxin toxicity on a decompensated heart-failure admission & 25 \\
\hline
3 & Family/Emergency Medicine & Rural ED ``triple whammy'' (NSAID+diuretic+ARB) AKI with salicylate toxicity masquerading as pneumonia/sepsis & 25 \\
\hline
4 & Emergency Physician & Ruptured abdominal aortic aneurysm at a district hospital with no vascular surgery; time-critical transfer decision & 30 \\
\hline
5 & Obstetric Registrar & Pre-eclampsia with severe asthma and twin pregnancy; labetalol relative-contraindication decision & 42 \\
\hline
\end{tabularx}

\subsection{3.2\quad Rubric schema}

Each rubric criterion is an atomic, MECE assertion about the ideal response, carrying: the criterion text; one or more guideline sources if appropriate; a justification; a weight from 1 (trivial) to 5 (critical, clinical-impact-weighted); one or more category tags (Reasoning, Safety, Extraction, Instruction Following, Style); an expert\_rating (true/false) per model; and three autorater fields (gpt\_autorater, claude\_autorater, gemini\_autorater).

\subsection{3.3\quad Inclusion threshold}

A task is retained only if the mean rubric score across the three frontier models is below 0.60. This targets the upper tail of clinical difficulty and ensures signal density (analogous in spirit to HealthBench Hard's selection of low-mean-score examples). All five tasks reported here satisfy this threshold (three-model means of 0.38, 0.57, 0.53, 0.48 and 0.09; Table 2).

\subsection{3.4\quad Quality-control pipeline}

\textbf{\textit{Phase 1: Structural QC.}} The three per-model rubrics for a task are verified to be identical except for expert\_rating.

\textbf{\textit{Phase 1.5: Peer review.}} A second clinician independently reviews each rubric for clinical defensibility, weight calibration, and missing criteria, converting single-expert opinion into specialty consensus.

\textbf{\textit{Phase 2: Autorater grading.}} Each of the three frontier models grades each model response against the rubric, producing three autorater labels per criterion alongside the expert label.

\textbf{\textit{Phase 2.5: Expert reconciliation.}} Criteria where $\geq$2 of 3 autoraters disagree with the expert are routed back to the clinician for re-adjudication; the expert's rating is final.

\textbf{\textit{Phase 3: Final consistency check.}} Structural identity is re-verified post-reconciliation.

\subsection{3.5\quad Models and harness}

We evaluated GPT 5.4 (OpenAI), Claude Opus 4.7 (Anthropic), and Gemini 3.1 Pro (Google DeepMind), all via API, single-turn. The rubric pass rate is the sum of weights of criteria with expert\_rating=true divided by the sum of all weights.

\subsection{3.6\quad Evaluation validity (third-party-evaluation framing)}

We adopt the vocabulary of OpenAI's third-party-evaluation guidance (OpenAI, 2026), which recommends that reports state the claim type, describe the harness, and account for validity hazards.

\begin{itemize}
\item \textbf{Claim type -- controlled comparison.} This study makes a controlled-comparison claim: three frontier models evaluated under a shared, fixed evaluation setup (identical tasks, rubrics, scoring, and harness).
\item \textbf{Harness.} The harness is a single-turn prompt to response package via API, with no agentic scaffolding, no additional tool access, no retrieval augmentation, and no retry logic. This is a deliberately conservative elicitation setup; models equipped with reasoning scaffolds, tools, or retrieval might perform differently on specific criteria.
\item \textbf{Contamination.} Scenarios are synthetic and clinician-authored, not drawn from any published case report or public dataset, so contamination risk is minimal.
\item \textbf{Refusals.} No model refused or partially refused any prompt in this pilot; no samples were compromised by refusals.
\item \textbf{Reward hacking.} Rubrics are expert-judged rather than automatically scored, and the peer-review and expert-reconciliation stages are designed to catch surface-correct-but-criterion-failing outputs (e.g., fluent responses that omit the decisive inference).
\item \textbf{Sandbagging.} Out of scope for a controlled-comparison capability study; we did not construct evaluation-awareness conditions.
\end{itemize}

\section{4.\quad Results}

\subsection{4.1\quad Aggregate scores}

\vspace{0.3em}
\noindent\textbf{Table 2. Rubric pass rate by task (weighted).}
\vspace{0.3em}

\noindent\begin{tabular}{|c|c|c|c|}
\hline
\textbf{Task} & \textbf{Gemini} & \textbf{GPT} & \textbf{Claude} \\
\hline
1 & 0.29 & 0.37 & 0.49 \\
\hline
2 & 0.60 & 0.46 & 0.64 \\
\hline
3 & 0.47 & 0.45 & 0.65 \\
\hline
4 & 0.42 & 0.57 & 0.45 \\
\hline
5 & 0.07 & 0.07 & 0.12 \\
\hline
\textbf{Mean} & \textbf{0.37} & \textbf{0.38} & \textbf{0.47} \\
\hline
\end{tabular}

\vspace{0.6em}

Claude led on 4 of 5 tasks and GPT on 1 of 5. Task 5 (which reflects pre-eclampsia with asthma and twin pregnancy) was a near-total failure for all three models (0.07--0.12). Per-task variance is high, consistent with a five-task sample.

\subsection{4.2\quad Failure distribution by weight class}

\vspace{0.3em}
\noindent\textbf{Table 3. Expert pass rate by criterion weight (aggregated across all five tasks).}
\vspace{0.3em}

\noindent\begin{tabular}{|c|c|c|c|}
\hline
\textbf{Weight} & \textbf{Gemini} & \textbf{GPT} & \textbf{Claude} \\
\hline
1 (trivial) & 8/10 = 80.0\% & 8/10 = 80.0\% & 9/10 = 90.0\% \\
\hline
2 & 1/1 = 100\% & 1/1 = 100\% & 1/1 = 100\% \\
\hline
3 & 12/44 = 27.3\% & 15/44 = 34.1\% & 20/44 = 45.5\% \\
\hline
4 & 5/21 = 23.8\% & 8/21 = 38.1\% & 8/21 = 38.1\% \\
\hline
5 (critical) & 35/108 = 32.4\% & 35/108 = 32.4\% & 45/108 = 41.7\% \\
\hline
\end{tabular}

\vspace{0.6em}

This is the central finding. Performance is \textit{inverted} relative to clinical priority: trivial weight-1 criteria pass at 80--90\%, while critical weight-5 criteria pass at only 32.4\% (Gemini and GPT) to 41.7\% (Claude). The criteria a clinician would consider most important (e.g., the ones whose omission would harm a patient) are the ones models most often miss.

\subsection{4.3\quad Failure distribution by category}

\vspace{0.3em}
\noindent\textbf{Table 4. Expert pass rate by category (aggregated; count-based).}
\vspace{0.3em}

\noindent\begin{tabular}{|c|c|c|c|}
\hline
\textbf{Category} & \textbf{Gemini} & \textbf{GPT} & \textbf{Claude} \\
\hline
Reasoning & 35/123 = 28.5\% & 38/123 = 30.9\% & 52/123 = 42.3\% \\
\hline
Safety & 26/67 = 38.8\% & 27/67 = 40.3\% & 30/67 = 44.8\% \\
\hline
Extraction & 1/3 = 33.3\% & 1/3 = 33.3\% & 1/3 = 33.3\% \\
\hline
Instruction Following & 7/8 = 87.5\% & 7/8 = 87.5\% & 7/8 = 87.5\% \\
\hline
Style & 5/5 = 100\% & 5/5 = 100\% & 5/5 = 100\% \\
\hline
\end{tabular}

\vspace{0.6em}

Instruction following (87.5\%) and style (100\%) are essentially solved for all three models. Reasoning is the largest category (123 criteria) and the weakest; Claude leads Reasoning by roughly 14 points over Gemini and 11 over GPT, and Safety by $\sim$5--6 points. The contrast between near-perfect formatting compliance and sub-45\% clinical reasoning is the qualitative signature of these failures.

\subsection{4.4\quad Autorater--expert agreement}

Across 552 expert-labelled criteria, each graded by all three LLM autoraters, raw met/not-met agreement with the expert was: GPT autorater 512/552 = 92.8\%; Gemini autorater 517/552 = 93.7\%; Claude autorater 522/552 = 94.6\%. We report raw percentage agreement because that is the statistic our reconciliation pipeline produces; we do not report kappa or F1, and formal inter-rater statistics over a larger sample are future work.

We are deliberately careful in how we contextualise these numbers. HealthBench responses were evaluated by a model-based grader, which the authors report can match expert grading, exceeding the average physician in five of seven themes with a class-balanced macro-F1 of 0.709 (Arora et al., 2025). That macro-F1 has a 0.50 random floor and is therefore not directly comparable to our raw percentage agreement, whose chance floor depends on label prevalence. Our figures are more directly comparable to raw exact-agreement statistics such as the 52.6\% physician--physician agreement on an ordinal score reported by Mehandru et al. (2025). The appropriate reading is that our autoraters track binary expert met/not-met judgement closely enough to serve as a \textit{calibration mirror}, not as a replacement for expert adjudication, which is precisely why expert reconciliation, not autorater consensus, is final in our pipeline.

\subsection{4.5\quad Universal critical misses}

We define a \textit{universal weight-5 miss} as a critical criterion that all three models failed. The counts were: Task 1, 11/30; Task 2, 5/15; Task 3, 2/12; Task 4, 4/13; Task 5, 33/38 -- 55 of 108 (51\%) critical criteria that no frontier model satisfied. Task 5 dominates: 33 of its 38 critical criteria were universally missed, which is why all three aggregate scores collapse on that task.

\subsection{4.6\quad Qualitative deep-dives: Examples of Universal misses (shared blind spots).}

\begin{itemize}
\item \textit{Example 1: Failure to use the improving BNP trend (1240 $\rightarrow$ 680) as evidence against ongoing volume overload.} All three models anchored on the admission heart-failure diagnosis and missed the contradicting biomarker trend. This is the canonical ``synthesis under contradiction'' failure: the decisive inference required overriding a salient prior with a later data point.
\item \textit{Example 2: Failure to identify the NSAID + furosemide + losartan ``triple whammy'' as the cause of acute kidney injury, and to explain that AKI reduces salicylate clearance and compounds toxicity.} The concurrent use of an NSAID, a diuretic, and a renin--angiotensin-system inhibitor produces an iatrogenic acute renal failure by simultaneously impairing glomerular autoregulation (through prostaglandin- and angiotensin-mediated effects) and impairing the natriuresis and increased renal blood flow expected with furosemide (Thomas, 2000). This impairment results in reduced salicylate elimination which increases the level and duration of salicylate-toxicity. No model connected the medication list to the AKI mechanism and onward to impaired drug clearance -- a two-step causal chain.
\item \textit{Example 3: Failure to state that labetalol is a relative contraindication in asthma.} This was the central safety criterion of the obstetric task, and all three models missed it. In this obstetric scenario, the central safety issue was not simply whether labetalol was ``contraindicated,'' but whether its benefits in managing uncontrolled pre-eclampsia outweighed the risk of bronchospasm in a patient with asthma. The models should have recognised labetalol as a first-line option in pre-eclampsia where clinically appropriate, while also recommending assessment of asthma severity and control before use. By presenting labetalol as simply contraindicated, they missed the required risk--benefit analysis. One study showed that 18.5\% of women with both pre-eclampsia and asthma received IV labetalol during delivery hospitalisations (Cabrera et al., 2018), supporting the case that clinicians balance risks when deciding treatment.
\item \textit{Example 4: Failure to state clearly that there should be no resuscitation in the event of loss of cardiac output during transfer.} In a ruptured AAA without on-site vascular surgery, cardiac arrest during transfer carries near-prohibitive mortality, and consensus guidance is that patients who arrest in the current episode should not be transferred. The safe action here is to \textit{withhold} an intervention -- a behaviour that completeness-rewarding evaluation does not naturally capture.
\end{itemize}

These examples are not cherry-picked superlatives; they are representative of the broader weight-5 distribution in Tables 3--4 and illustrate the kinds of inferences (synthesis under contradiction, mechanism reasoning, resource anticipation, and knowing when to withhold) that frontier models most reliably miss.

\section{5.\quad Discussion}

\textbf{Instruction-following is solved; clinical reasoning and safety are not.} The most actionable result is the dissociation in Table 4: models reliably produce correctly formatted, well-styled artefacts (87.5--100\%) while missing the majority of the reasoning and safety content those artefacts are supposed to carry (28.5--44.8\%). For anyone deploying these models in clinical workflows, the implication is that surface quality is a misleading proxy for clinical correctness -- a fluent, well-organised SOAP note can omit the single inference that matters. This mirrors the observation of Mehandru et al. (2025) that polished outputs can mask reasoning that degrades as evidence accumulates.

\textbf{Atomic weighted rubrics surface omission failures that preference ratings miss.} A pairwise or Likert preference comparison would likely rate all three models' responses as fluent and plausible. It is precisely the atomic, weighted, golden-answer-motivated rubric that exposes the weight-5 misses, because each decisive inference is its own criterion and is scored independently of the surrounding prose. The inversion in Table 3 is invisible to holistic scoring and visible to atomic scoring.

\textbf{Implications for training.} The discriminating criteria (universal misses or criteria where 2 models passed and 1 failed) are natural sources of supervision. Where one model passes a weight-5 criterion that a peer fails, the pair constitutes a ready-made contrast for preference optimisation (DPO) or a target for rubric-reward RL (Gunjal et al., 2025), with the weight serving as a partial-credit signal. Because the rubrics are expert-grounded and decomposed, they are less susceptible to the reward-hacking failure modes of opaque preference models.

\section{6.\quad Limitations}

This is a methods-and-preliminary-findings contribution, and we state its limits plainly. First, with only five tasks and 184 criteria we cannot conduct statistical inference about model rankings; the mean differences in Table 2 should be read as descriptive, not significant. The contribution is the pipeline and the qualitative failure taxonomy; the dataset is intended to develop into a large-scale benchmark with the statistical power this pilot lacks. Second, rubrics and scenarios cite predominantly UK/US guidance (RCoA/NAP6--7, Resuscitation Council UK, NICE, RCEM, SVS), which may not transfer to other health systems. Furthermore, each criterion has one expert label (peer-reviewed and reconciled where autoraters disagreed); we do not report independent double-expert grading on every criterion, and formal inter-rater statistics (kappa, macro-F1) over a larger sample are future work.

We frame the multi-stage, clinician-authored, peer-reviewed pipeline as the scalable contribution: the same process applied to hundreds of scenarios across specialties would yield a benchmark with the statistical power this pilot lacks, while preserving the validity guarantee that rubrics derive from clinician golden answers rather than model outputs.

\section{7.\quad Conclusion}

Across five clinician-authored, upper-tail-difficulty clinical scenarios, three frontier models passed only a minority of critical (weight-5) rubric criteria (32.4--41.7\%) while nearly saturating instruction-following and style -- an inversion of clinical priority in which the most important inferences are the most frequently missed. A total of 55 of 108 critical (weight-5) criteria (51\%) were satisfied by no model, concentrated in synthesis under contradictory evidence, mechanism reasoning, resource anticipation, and knowing when to withhold an intervention. Three LLM autoraters reproduced expert labels on 92.8--94.6\% of criteria, supporting their use as a calibration mirror. We release this as a demonstration of a scalable, validity-preserving evaluation pipeline, ready to develop into a large-scale benchmark, and we invite the ML4H and frontier-lab communities to extend it.

\section{References}

\refentry{Arora, R. K., Wei, J., Soskin Hicks, R., Bowman, P., Quiñonero-Candela, J., Tsimpourlas, F., Sharman, M., Shah, M., Vallone, A., Beutel, A., Heidecke, J., \& Singhal, K. (2025). \textit{HealthBench: Evaluating large language models towards improved human health.} arXiv. \url{https://arxiv.org/abs/2505.08775}}

\refentry{Bedi, S., Jain, S. S., Chandra, R., Pierson, E., Koyejo, S., Stoyanovich, J., \& Shah, N. H. (2025). \textit{MedHELM: Holistic evaluation of large language models for medical applications.} arXiv. \url{https://arxiv.org/abs/2505.23802}}

\refentry{Cabrera, C. I., Bui, T. L., \& Andrade, J. (2018). Use of antihypertensive medications and uterotonics during delivery hospitalizations in women with asthma. \textit{Obstetrics \& Gynecology, 131}(6), 1057--1064.}

\refentry{Gunjal, A., Wang, A., Lau, E., Nath, V., Liu, B., \& Hendryx, S. M. (2025). \textit{Rubrics as rewards: Reinforcement learning beyond verifiable domains.} arXiv. \url{https://arxiv.org/abs/2507.17746}}

\refentry{Jin, D., Pan, E., Oufattole, N., Weng, W.-H., Fang, H., \& Szolovits, P. (2021). What disease does this patient have? A large-scale open domain question answering dataset from medical exams. \textit{Applied Sciences, 11}(14), 6421.}

\refentry{Jin, Q., Dhingra, B., Liu, Z., Cohen, W., \& Lu, X. (2019). PubMedQA: A dataset for biomedical research question answering. In \textit{Proceedings of the 2019 Conference on Empirical Methods in Natural Language Processing.}}

\refentry{Liu, T., Xu, Z., Hu, Z., Shi, W., Zhuang, Y., \& Yu, H. (2025). \textit{OpenRubrics: Towards scalable synthetic rubric generation for reward modeling and LLM alignment.} arXiv. \url{https://arxiv.org/abs/2510.07743}}

\refentry{Mehandru, N., Golchini, N., Bamman, D., Zack, T., Molina, M. F., \& Alaa, A. (2025). \textit{ER-Reason: A benchmark dataset for LLM-based clinical reasoning in the emergency room.} arXiv. \url{https://arxiv.org/abs/2505.22919}}

\refentry{OpenAI. (2026). \textit{A shared playbook for trustworthy third-party evaluations.} \url{https://openai.com/index/trustworthy-third-party-evaluations-foundations/}}

\refentry{Pal, A., Umapathi, L. K., \& Sankarasubbu, M. (2022). MedMCQA: A large-scale multi-subject multi-choice dataset for medical domain question answering. In \textit{Proceedings of the Conference on Health, Inference, and Learning.}}

\refentry{Palepu, A., Schaekermann, M., Tu, T., Palfia, B., Saab, K., Tanno, R., Barral, J., Karthikesalingam, A., \& Natarajan, V. (2025). \textit{Towards conversational AI for disease management.} arXiv. \url{https://arxiv.org/abs/2503.06074}}

\refentry{Saab, K., Tanno, R., Palepu, A., Tu, T., Schaekermann, M., Barral, J., Webster, D., Karthikesalingam, A., \& Natarajan, V. (2025). \textit{Advancing conversational diagnostic AI with multimodal reasoning.} arXiv. \url{https://arxiv.org/abs/2505.04653}}

\refentry{Singhal, K., Azizi, S., Tu, T., Mahdavi, S. S., Wei, J., Chung, H. W., Scales, N., Tanwani, A., Cole-Lewis, H., Pfohl, S., Payne, P., Seneviratne, M., Gamble, P., Kelly, C., Babiker, A., Schärli, N., Chowdhery, A., Mansfield, P., Demner-Fushman, D., \ldots Natarajan, V. (2023). Large language models encode clinical knowledge. \textit{Nature, 620}(7972), 172--180.}

\refentry{Singhal, K., Tu, T., Gottweis, J., Sayres, R., Wulczyn, E., Amin, M., Hou, L., Clark, K., Pfohl, S. R., Cole-Lewis, H., Neal, D., Rashid, Q. M., Schaekermann, M., Wang, A., Dash, D., Chen, J. H., Shah, N. H., Lachgar, S., Mansfield, P., \ldots Natarajan, V. (2025). Toward expert-level medical question answering with large language models. \textit{Nature Medicine, 31}, 943--950.}

\refentry{Thomas, M. C. (2000). Diuretics, ACE inhibitors and NSAIDs -- the triple whammy. \textit{The Medical Journal of Australia, 172}(4), 184--185.}

\refentry{Tu, T., Schaekermann, M., Palepu, A., Saab, K., Freyberg, J., Tanno, R., Wang, A., Li, B., Amin, M., Cheng, Y., Vedadi, E., Tomasev, N., Azizi, S., Singhal, K., Hou, L., Webson, A., Kulkarni, K., Mahdavi, S. S., Semturs, C., \ldots Natarajan, V. (2025). Towards conversational diagnostic artificial intelligence. \textit{Nature, 642}(8067), 442--450.}

\refentry{Zheng, L., Chiang, W.-L., Sheng, Y., Zhuang, S., Wu, Z., Zhuang, Y., Lin, Z., Li, Z., Li, D., Xing, E. P., Zhang, H., Gonzalez, J. E., \& Stoica, I. (2023). Judging LLM-as-a-judge with MT-Bench and Chatbot Arena. In \textit{Advances in Neural Information Processing Systems 36.}}

\end{document}